%% file: main.tex
\documentclass[10pt,twocolumn,letterpaper]{article}

\usepackage[accsupp]{axessibility}  % Improves PDF readability for those with disabilities.

\usepackage{cvpr}              % To produce the CAMERA-READY version
\input{preamble}

\usepackage{marvosym}

\definecolor{cvprblue}{rgb}{0.21,0.49,0.74}
\usepackage[pagebackref,breaklinks,colorlinks,allcolors=cvprblue]{hyperref}

\def\paperID{8342} % *** Enter the Paper ID here
\def\confName{CVPR}
\def\confYear{2025}

\title{CarPlanner: Consistent Auto-regressive Trajectory Planning for Large-scale Reinforcement Learning in Autonomous Driving}

\newcommand*{\affaddr}[1]{#1} % No op here. Customize it for different styles.
\newcommand*{\affmark}[1][*]{\textsuperscript{#1}}

\author{
Dongkun Zhang$^{1,2}$\quad Jiaming Liang$^{2}$\quad Ke Guo$^{2}$\quad Sha Lu$^{1}$\quad Qi Wang$^{2}$ \\ Rong Xiong$^{1,\textrm{\Letter}}$\quad Zhenwei Miao$^{2,\dag}$\quad Yue Wang$^{1}$ \\
\affaddr{\affmark[1]Zhejiang University}\quad
\affaddr{\affmark[2]Cainiao Network}\\
{\affmark[1] \tt\small \{zhangdongkun, lusha, rxiong, ywang24\}@zju.edu.cn} \\
{\affmark[2] \tt\small \{liangjiaming.ljm, muguo.gk, ruifeng.wq, zhenwei.mzw\}@alibaba-inc.com}
}

\begin{document}
\maketitle
\def\thefootnote{\dag}\footnotetext[1]{Project lead. $^\textrm{\Letter}$ Corresponding author.}

\input{sections/0-abstract}

\input{sections/1-intro}

\input{sections/2-related-work}

\input{sections/3-method}

\input{sections/4-results}

\input{sections/5-conclusion}

{
    \small
    \bibliographystyle{ieeenat_fullname}
    \bibliography{main}
}

\input{sections/6-supple}

\end{document}

%% file: preamble.tex
\usepackage{makecell}
\usepackage{multirow}
\usepackage{colortbl}

\usepackage{pifont} % http://ctan.org/pkg/pifont

\newcommand{\cmark}{\textcolor{green}{\ding{51}}}
\newcommand{\xmark}{\textcolor{red}{\ding{55}}}

\usepackage{algorithmicx}
\usepackage{algorithm}
\usepackage{algpseudocode}
\usepackage{listings}

\definecolor{DeepSkyBlue}{RGB}{0,191,255}
\definecolor{Blue}{RGB}{31,119,180}

%% file: sections/0-abstract.tex
\begin{abstract}
Trajectory planning is vital for autonomous driving, ensuring safe and efficient navigation in complex environments. While recent learning-based methods, particularly reinforcement learning (RL), have shown promise in specific scenarios, RL planners struggle with training inefficiencies and managing large-scale, real-world driving scenarios.
In this paper, we introduce \textbf{CarPlanner}, a \textbf{C}onsistent \textbf{a}uto-\textbf{r}egressive \textbf{Planner} that uses RL to generate multi-modal trajectories. The auto-regressive structure enables efficient large-scale RL training, while the incorporation of consistency ensures stable policy learning by maintaining coherent temporal consistency across time steps. Moreover, CarPlanner employs a generation-selection framework with an expert-guided reward function and an invariant-view module, simplifying RL training and enhancing policy performance.
Extensive analysis demonstrates that our proposed RL framework effectively addresses the challenges of training efficiency and performance enhancement, positioning CarPlanner as a promising solution for trajectory planning in autonomous driving.
To the best of our knowledge, we are the first to demonstrate that the RL-based planner can surpass both IL- and rule-based state-of-the-arts (SOTAs) on the challenging large-scale real-world dataset nuPlan. Our proposed CarPlanner surpasses  RL-, IL-, and rule-based SOTA approaches within this demanding dataset.
\end{abstract}

%% file: sections/1-intro.tex
\section{Introduction}

\begin{figure}[t]
    \begin{center}
    \includegraphics[width=0.44\textwidth]{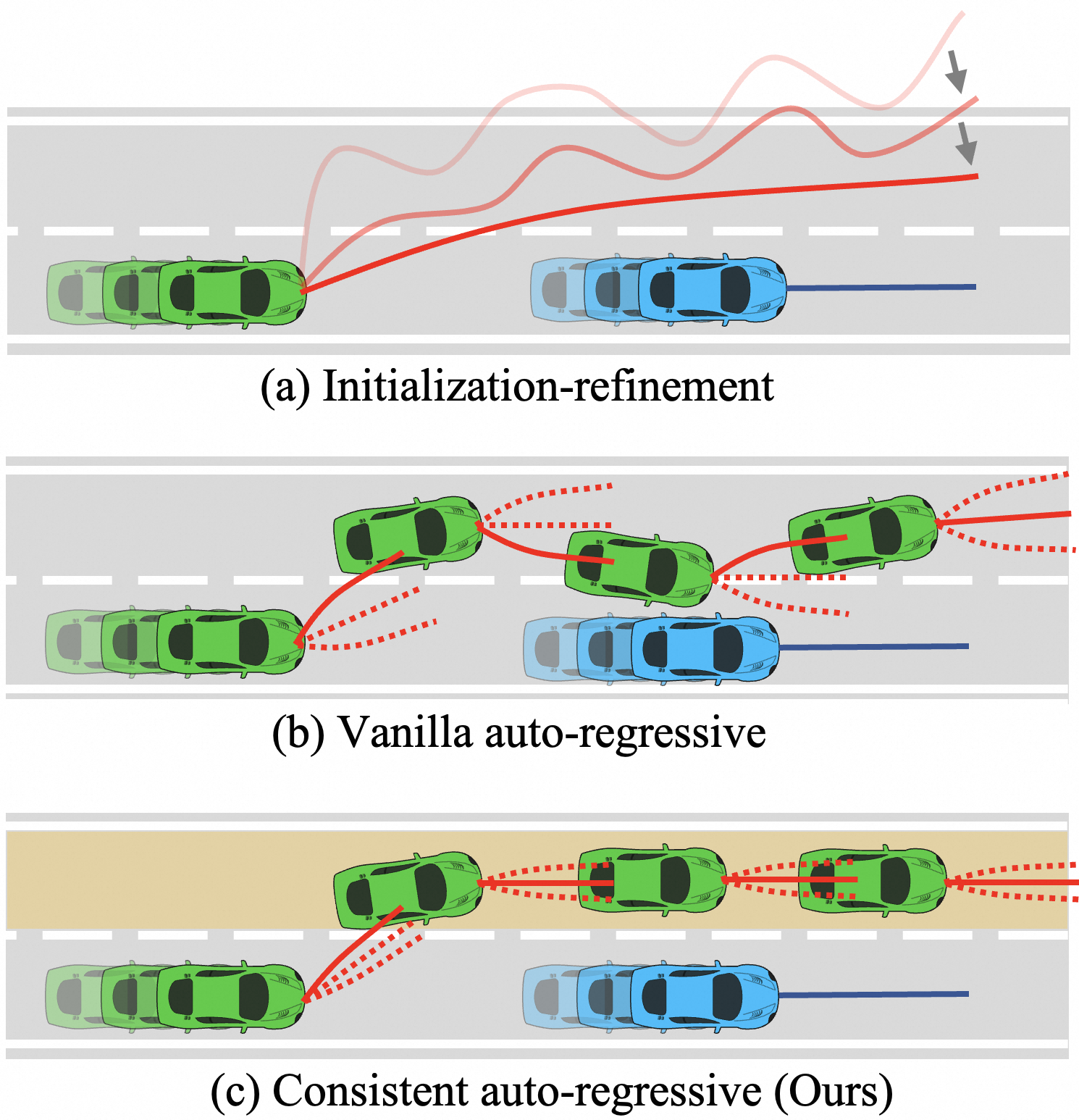}
    %%% trim={<left> <lower> <right> <upper>}
    \end{center}
    \vspace{-0.5cm}
    \caption{Frameworks for multi-step trajectory generation. (a)~Initialization-refinement that generates an initial trajectory and refines it iteratively. (b)~Vanilla auto-regressive models that decode subsequent poses sequentially. (c)~Our consistent auto-regressive model that integrates time-consistent mode information.
    }
    \vspace{-0.6cm}
    \label{figure:teaser}
\end{figure}

Trajectory planning~\cite{tampuu2020survey} is essential in autonomous driving, utilizing outputs from perception and trajectory prediction modules to generate future poses for the ego vehicle. A controller tracks this planned trajectory, producing control commands for closed-loop driving. Recently, learning-based trajectory planning has garnered attention due to its potential to automate algorithm iteration, eliminate tedious rule design, and ensure safety and comfort in diverse real-world scenarios~\cite{tampuu2020survey}. 

Most existing researches~\cite{scheel2022urban,Guo-RSS-23,hu2023planning} employ imitation learning~(IL) to align planned trajectories with those of human experts. However, this approach suffers from distribution shift~\cite{ross2011reduction} and causal confusion~\cite{de2019causal}. Reinforcement learning~(RL) offers a potential solution, addressing these challenges and providing richer supervision through reward functions. {Although RL shows effectiveness in domains such as games~\cite{silver2016mastering}, robotics~\cite{ibarz2021train}, and language models~\cite{ouyang2022training}, it still struggles with training inefficiencies and performance issues in the large-scale driving task.
}
To the extent of our knowledge, no RL methods have yet achieved competitive results on large-scale open datasets such as nuPlan~\cite{caesar2021nuplan}, which features diverse real-world scenarios.

Thus, this paper aims to tackle two key challenges in RL for trajectory planning: 1) training inefficiency and 2) poor performance. Training inefficiency arises from the fact that RL typically operates in a model-free setting, necessitating an inefficient simulator running on a CPU to repeatedly roll out a policy for data collection. To overcome this challenge, we propose an efficient model-based approach utilizing neural networks as transition models. Our method is optimized for execution on hardware accelerators such as GPUs, rendering our time cost comparable to that of IL-based methods.

To apply RL to solve the trajectory planning problem, we formulate it as a multi-step sequential decision-making task utilizing a Markov Decision Process~(MDP). Existing methods that generate the trajectory\footnote{In this paper, the term ``trajectory'' refers to the future poses of the ego vehicle or traffic agents. To avoid confusion, we use the term ``state (action) sequence'' to refer to the ``trajectory'' in the RL community.} in multiple steps generally fall into two categories: initialization-refinement~\cite{shi2022motion,huang2023gameformer, jiang2023motiondiffuser,zhou2024smartrefine} and auto-regressive models~\cite{rhinehart2019precog,seff2023motionlm,zhou2024behaviorgpt,wu2024smart}.

The first category, illustrated in \cref{figure:teaser} (a), involves generating an initial trajectory estimate and subsequently refining it through iterative applications of RL. However, recent studies, including Gen-Drive~\cite{huang2024gen}, suggest that it continues to lag behind SOTA IL and rule-based planners. 
One notable limitation of this approach is its neglect of the temporal causality inherent in the trajectory planning task. Additionally, the complexity of {direct optimization over high-dimensional trajectory space} can hinder the performance of RL algorithms.
The second category consists of auto-regressive models, shown in \cref{figure:teaser} (b), which generate the poses of the ego vehicle recurrently using a single-step policy within a transition model. In this category, ego poses at all time steps are consolidated to form the overall planned trajectory. {As taking temporal causality into account, current auto-regressive models allow for interactive behaviors}. However, a common limitation is their reliance on auto-regressively random sampling from action distributions to generate multi-modal trajectories. This vanilla auto-regressive procedure may compromise long-term consistency and {unnecessarily} expand the exploration space in RL, leading to poor performance.

To address the limitations of auto-regressive models, we introduce \textbf{CarPlanner}, a \textbf{C}onsistent \textbf{a}uto-\textbf{r}egressive model designed for efficient, large-scale RL-based \textbf{Planner} training (see \cref{figure:teaser} (c)). The key insight of CarPlanner is its incorporation of consistent mode representation as conditions for the auto-regressive model. Specifically, we leverage a longitudinal-lateral decomposed mode representation, where the longitudinal mode is a scalar that captures average speeds, and the lateral mode encompasses all possible routes derived from the current state of the ego vehicle along with map information. This mode remains constant across time steps, providing stable and consistent guidance during policy sampling.

{Furthermore, we propose a universal reward function that suits large-scale and diverse scenarios, eliminating the need for scenario-specific reward designs. This function consists of an expert-guided and task-oriented term.
The first term quantifies the displacement error between the ego-planned trajectory and the expert's trajectory, which, along with the consistent mode representation, narrows down the policy's exploration space.
The second term incorporates common senses in driving tasks including the avoidance of collision and adherence to the drivable area.
Additionally, we introduce an Invariant-View Module~(IVM) to supply invariant-view input for policy, with the aim of providing time-agnostic policy input, easing the feature learning and embracing generalization. To achieve this, IVM preprocesses state and lateral mode by transforming agent, map, and route information into the ego's current coordinate and by clipping information that is distant from the ego.}

\textit{\textbf{To our knowledge, we are the first to demonstrate that RL-based planner outperforms state-of-the-art~(SOTA) IL and rule-based approaches on the challenging large-scale nuPlan dataset}}. In summary, the key contributions of this paper are highlighted as follows:
\begin{itemize}
    \item We present \textbf{CarPlanner}, a consistent auto-regressive planner that trains an RL policy to generate consistent multi-modal trajectories.
    \item We introduce an expert-guided universal reward function and IVM to simplify RL training and improve policy generalization, {leading to enhanced closed-loop performance}.
    \item We conduct a rigorous analysis on the characteristics of IL and RL training, providing insights into their strengths and limitations, while highlighting the advantages of RL in tackling challenges such as distribution shift and causal confusion.
    \item {Our framework showcases exceptional performance, surpassing all  RL-, IL-, and rule-based SOTAs on the nuPlan benchmark. This underscores the potential of RL in navigating complex real-world driving scenarios.}
\end{itemize}

%% file: sections/2-related-work.tex
\section{Related Work}

\begin{figure*}[t]
    \begin{center}
    \includegraphics[width=0.95\textwidth]{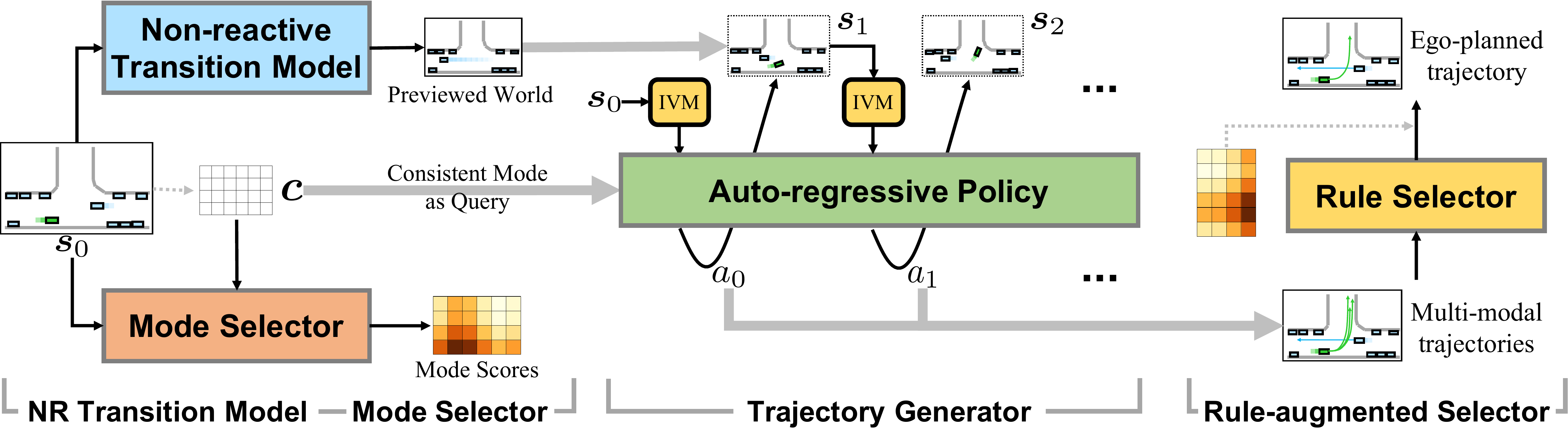}
    %%% trim={<left> <lower> <right> <upper>}
    \end{center}
    \vspace{-0.6cm}
    \caption{CarPlanner contains four parts. (1) The non-reactive transition model takes initial state $\boldsymbol{s}_0$ as input and predicts the future trajectories of traffic agents. (2) The mode selector outputs scores based on the initial state and the modes $\boldsymbol{c}$. (3) The trajectory generator obeys an auto-regressive structure condition on the consistent mode and produces mode-aligned multi-modal trajectories. (4) The rule-augmented selector compensates the mode scores by safety, comfort, and progress metrics.
    }
    \vspace{-0.5cm}
    \label{figure:planner}
\end{figure*}

\subsection{Imitation-based Planning}
The use of IL to train planners based on human demonstrations has garnered significant interest recently. This approach leverages the driving expertise of experienced drivers who can safely and comfortably navigate a wide range of real-world scenarios, along with the added advantage of easily collectible driving data at scale~\cite{ettinger2021large, houston2021one, caesar2021nuplan}. Numerous studies~\cite{chitta2022transfuser, scheel2022urban, huang2023gameformer} have focused on developing innovative networks to enhance open-loop performance in this domain. However, the ultimate challenge of autonomous driving is achieving closed-loop operation, which is evaluated using driving-oriented metrics such as safety, adherence to traffic rules, comfort, and progress. This reveals a significant gap between the training and testing phases of planners. Moreover, IL is particularly vulnerable to issues such as distribution shift~\cite{ross2011reduction} and causal confusion~\cite{de2019causal}. The first issue results in suboptimal decisions when the system encounters scenarios that are not represented in the training data distribution. The second issue arises when networks inadvertently capture incorrect correlations and develop shortcuts based on input information, primarily due to the reliance on imitation loss from expert demonstrations. Despite efforts in several studies~\cite{Ogale-RSS-19, cheng2024rethinking, zhang2024pep, cheng2024pluto} to address these challenges, the gap between training and testing remains substantial.

\subsection{RL in Autonomous Driving}
In the field of autonomous driving, RL has demonstrated its effectiveness in addressing specific scenarios such as highway driving~\cite{leurent2019social, wang2021deep}, lane changes~\cite{li2022decision, he2022robust}, and unprotected left turns~\cite{leurent2019social, Wang-RSS-23}. Most methods directly learn policies over the control space, which includes throttle, brake, and steering commands. Due to the high frequency of control command execution, the simulation process can be time-consuming, and exploration can be inconsistent~\cite{Wang-RSS-23}. Several works~\cite{zhou2023accelerating, Wang-RSS-23} have proposed learning trajectory planners with actions defined as ego-planned trajectories, which temporally extend the exploration space and improve training efficiency. However, a trade-off exists between the trajectory horizon and training performance, as noted in ASAP-RL~\cite{Wang-RSS-23}. Increasing the trajectory horizon results in less reactive behaviors and a reduced amount of data, while a smaller trajectory horizon leads to challenges similar to those encountered in control space. Additionally, these methods typically employ a model-free setting, making them difficult to apply to the complex, diverse real-world scenarios found in large-scale driving datasets. In this paper, we propose adopting a model-based formulation that can facilitate RL training on large-scale datasets. Under this formulation, we aim to overcome the trajectory horizon trade-off by using a transition model, which can provide a preview of the world in which our policy can make multi-step decisions during testing.

%% file: sections/3-method.tex
\section{Method}

\subsection{Preliminaries}

\textbf{MDP} is used to model sequential decision problems, formulated as a tuple $\left<\mathcal{S}, \mathcal{A}, P_{\tau}, R, \rho_0, \gamma, T\right>$.
$\mathcal{S}$ is the state space.
$\mathcal{A}$ is the action space.
$P_{\tau} : \mathcal{S} \times \mathcal{A} \to \Delta(\mathcal{S})$ \footnote{$\Delta(\mathcal{X})$ denotes the set of probability distribution over set $\mathcal{X}$.} is the state transition probability.
$R : \mathcal{S} \times \mathcal{A} \to \mathbb{R}$  denotes the reward function and is bounded.
$\rho_0 \in \Delta(\mathcal{S})$ is the initial state distribution.
$T$ is the time horizon and $\gamma$ is the discount factor of future rewards.
The state-action sequence is defined as $\tau = (\boldsymbol{s}_0, a_0, \boldsymbol{s}_1, a_1, \dots, \boldsymbol{s}_T)$, where $\boldsymbol{s}_t \in \mathcal{S}$ and $a_t \in \mathcal{A}$ are the state and action at time step $t$.
The objective of RL is to maximize the expected return:
\begin{equation}
\setlength{\abovedisplayskip}{0pt}
\begin{split}
    \max_{\pi} \mathbb{E}_{\boldsymbol{s}_t \sim {P_{\tau}}, a_t \sim \pi} [\sum_{t=0}^{T}\gamma^{t} R(\boldsymbol{s}_t, a_t)].  \label{equation:mdp}
\end{split}
\setlength{\belowdisplayskip}{0pt}
\end{equation}

\noindent\textbf{Vectorized state representation}. State $\boldsymbol{s}_t$ contains map and agent information in vectorized representation~\cite{gao2020vectornet}.
Map information $m$ includes the road network, traffic lights, etc, which are represented by polylines and polygons.
Agent information includes the current and past poses of ego vehicle and other traffic agents, which are represented by polylines. The index of ego vehicle is $0$ and the indices of traffic agents range from $1$ to $N$. For each agent $i$, its history is denoted as $s^{i}_{t-H:t}, i \in \{0, 1, \dots, N\}$, where $H$ is the history time horizon.

\subsection{Problem Formulation}

We model the trajectory planning task as a sequential decision process and decouple the auto-regressive models into policy and transition models.
The key to connect trajectory planning and auto-regressive models is to define the action as the next pose of ego vehicle, i.e., $a_t = s^{0}_{t+1}$. Therefore, after forwarding the auto-regressive model, the decoded pose is collected to be the ego-planned trajectory.
Specifically, we can reduce the state-action sequence to the state sequence under this definition and vectorized representation:
\begin{equation}
\setlength{\abovedisplayskip}{0pt}
\begin{split}
    &P(\boldsymbol{s}_0, a_0, \boldsymbol{s}_1, a_1, \dots, \boldsymbol{s}_T) \\
    &= P(m, s^{0:N}_{-H:0}, s^{0}_{1}, m, s^{0:N}_{1-H:1}, s^{0}_{2}, \dots, m, s^{0:N}_{T-H:T}) \\
    &= P(m, s^{0:N}_{-H:0}, m, s^{0:N}_{1-H:1}, \dots, m, s^{0:N}_{T-H:T}) \\
    &= P(\boldsymbol{s}_0, \boldsymbol{s}_1, \dots, \boldsymbol{s}_T).
    \\[-0.1cm]
\end{split}
\setlength{\belowdisplayskip}{0pt}
\end{equation}
The state sequence can be further formulated in an auto-regressive fashion and decomposed into policy and transition model:
\begin{equation}
\setlength{\abovedisplayskip}{0pt}
\begin{split}
    &P(\boldsymbol{s}_0, \boldsymbol{s}_1, \dots, \boldsymbol{s}_T) = \rho_{0}(\boldsymbol{s}_0) \prod_{t=0}^{T-1} P(\boldsymbol{s}_{t+1} | \boldsymbol{s}_t) \\
    &= \rho_{0}(\boldsymbol{s}_0) \prod_{t=0}^{T-1} P(s^0_{t+1}, s^{1:N}_{t+1} | \boldsymbol{s}_t) \\
    &= \rho_{0}(\boldsymbol{s}_0) \prod_{t=0}^{T-1} \underbrace{\pi(a_t | \boldsymbol{s}_t)}_{\text{Policy}} \underbrace{P_{\tau}(s^{1:N}_{t+1} | \boldsymbol{s}_t)}_{\text{Transition Model}}.
    \\[-0.15cm]
\end{split}
\label{equation:incons}
\setlength{\belowdisplayskip}{0pt}
\end{equation}
From \cref{equation:incons}, we can clearly identify the inherent problem associated with the typical auto-regressive approach: inconsistent behaviors across time steps arise from the policy distribution, which depends on random sampling from the action distribution.

To solve the above problem, we introduce consistent mode information $\boldsymbol{c}$ that remains unchanged across time steps into the auto-regressive fashion:
\begin{equation}
\setlength{\abovedisplayskip}{0pt}
\begin{split}
&P(\boldsymbol{s}_0, \boldsymbol{s}_1, \dots, \boldsymbol{s}_T) = \int_{\boldsymbol{c}} P(\boldsymbol{s}_0, \boldsymbol{s}_1, \dots, \boldsymbol{s}_T, \boldsymbol{c}) d\boldsymbol{c}  \\
&= \rho_0(\boldsymbol{s}_0) \int_{\boldsymbol{c}} P(\boldsymbol{c} | \boldsymbol{s}_0) P(\boldsymbol{s}_1, \dots, \boldsymbol{s}_T | \boldsymbol{c}) d\boldsymbol{c} \\
&= \rho_0(\boldsymbol{s}_0) \prod_{t=0}^{T-1} \underbrace{P_{\tau}(s^{1:N}_{t+1} | \boldsymbol{s}_t)}_{\text{Transition Model}} \int_{\boldsymbol{c}} \underbrace{P(\boldsymbol{c} | \boldsymbol{s}_0)}_{\text{Mode Selector}} \prod_{t=0}^{T-1} \underbrace{\pi(a_t | \boldsymbol{s}_t, \boldsymbol{c})}_{\text{Policy}}   d\boldsymbol{c}. 
\\[-1.cm]
\label{equation:car}
\end{split}
\setlength{\belowdisplayskip}{0pt}
\end{equation}
Since we focus on the ego trajectory planning, the consistent mode $\boldsymbol{c}$ does not impact transition model.

This consistent auto-regressive formulation defined in \cref{equation:car} reveals a generation-selection framework where the mode selector scores each mode based on the initial state $s_0$ and the trajectory generator generates multi-modal trajectories via sampling from the mode-conditioned policy.

\noindent\textbf{Non-reactive transition model.} The transition model formulated in \cref{equation:car} {needs to be employed in every time step since it} produces the poses of traffic agents at time step $t+1$ based on current state $\boldsymbol{s}_t$. In practice, this process is time-consuming and we do not observe a performance improvement by using this transition model, therefore, we use trajectory predictors $P(s^{1:N}_{1:T} | \boldsymbol{s}_0)$ as non-reactive transition model {that produces all future poses of traffic agents in one shot given initial state $\boldsymbol{s}_0$}.

\subsection{Planner Architecture}

The framework of our proposed \textbf{CarPlanner} is illustrated in \cref{figure:planner}, comprising four key components: 1) the non-reactive transition model, 2) the mode selector, 3) the trajectory generator, and 4) the rule-augmented selector.

Our planner operates within a generation-selection framework. Given an initial state $s_0$ and all possible  $N_{\text{mode}}$ modes, the trajectory selector evaluates and assigns scores to each mode. The trajectory generator then produces $N_{\text{mode}}$ trajectories that correspond to their respective modes.
For trajectory generator, the initial state $s_0$ is replicated $N_{\text{mode}}$ times, each associated with one of the $N_{\text{mode}}$ modes, effectively creating $N_{\text{mode}}$ parallel worlds. The policy is executed within these previewed worlds. During the policy rollout, a trajectory predictor acts as the state transition model, generating future poses of traffic agents across all time horizons.

\subsubsection{Non-reactive Transition Model}

This module takes the initial state $s_0$ as input and outputs the future trajectories of traffic agents. The initial state is processed by agent and map encoders, followed by a self-attention Transformer encoder~\cite{vaswani2017attention} to fuse the agent and map features. The agent features are then decoded into future trajectories.

\noindent\textbf{Agent and map encoders.} The state $s_0$ contains both map and agent information. The map information $m$ consists of $N_{m,1}$ polylines and $N_{m,2}$ polygons. The polylines describe lane centers and lane boundaries, with each polyline containing $3N_p$ points, where $3$ corresponds to the lane center, the left boundary, and the right boundary. Each point is with dimension $D_m=9$ and includes the following attributes: x, y, heading, speed limit, and category. When concatenated, the points of the left and right boundaries together with the center point yield a dimension of $N_{m,1} \times N_p \times 3D_m$. We leverage a PointNet~\cite{qi2017pointnet} to extract features from the points of each polyline, resulting in a dimensionality of $N_{m,1} \times D$, where $D$ represents the feature dimension. The polygons represent intersections, crosswalks, stop lines, etc, with each polygon containing $N_p$ points. We utilize another PointNet to extract features from the points of each polygon, producing a dimension of $N_{m,2} \times D$. We then concatenate the features from both polylines and polygons to form the overall map features, resulting in a dimension of $N_{m} \times D$. The agent information $A$ consists of $N$ agents, where each agent maintains poses for the past $H$ time steps. Each pose is with dimension $D_a=10$ and includes the following attributes: x, y, heading, velocity, bounding box, time step, and category. Consequently, the agent information has a dimension of $N \times H \times D_a$. We apply another PointNet to extract features from the poses of each agent, yielding an agent feature dimension of $N \times D$.

\subsubsection{Mode Selector}
This module takes \( s_0 \) and longitudinal-lateral decomposed mode information as input and outputs the probability of each mode. The number of modes $N_{\text{mode}} = N_{\text{lat}} N_{\text{lon}}$.

\noindent\textbf{Route-speed decomposed mode.} To capture the longitudinal behaviors, we generate \( N_{\text{lon}} \) modes that represent the average speed of the trajectory associated with each mode. Each longitudinal mode \( c_{\text{lon},j} \) is defined as a scalar value of \( \frac{j}{N_{\text{lon}}} \), repeated across a dimension \( D \). As a result, the dimensionality of the longitudinal modes is \( N_{\text{lon}} \times D \). For lateral behaviors, we identify \( N_{\text{lat}} \) possible routes from the map using a graph search algorithm. These routes correspond to the lanes available for the ego vehicle. The dimensionality of these routes is \( N_{\text{lat}} \times N_r \times D_m \). We employ another PointNet to aggregate the features of the \( N_r \) points along each route, producing a lateral mode with a dimension of \( N_{\text{lat}} \times D \). To create a comprehensive mode representation $\boldsymbol{c}$, we combine the lateral and longitudinal modes, resulting in a combined dimension of \( N_{\text{lat}} \times N_{\text{lon}} \times 2D \). To align this mode information with other feature dimensions, we pass it through a linear layer, mapping it back to \( N_{\text{lat}} \times N_{\text{lon}} \times D \).

\noindent\textbf{Query-based Transformer decoder.} This decoder is employed to fuse the mode features with map and agent features derived from \( s_0 \). In this framework, the mode serves as the query, while the map and agent information act as the keys and values. The updated mode features are decoded through a multi-layer perceptron (MLP) to yield the scores for each mode, which are subsequently normalized using the softmax operator.

\subsubsection{Trajectory Generator}
This module operates in an auto-regressive manner, recurrently decoding the next pose of the ego vehicle $a_t$, given the current state $\boldsymbol{s}_t$, and consistent mode information $\boldsymbol{c}$.

\noindent\textbf{Invariant-view module (IVM).} Before feeding the mode and state into the network, we preprocess them to eliminate time information. For the map and agent information in state $\boldsymbol{s}_t$, we select the $K$-nearest neighbors~(KNN) to the ego current pose and only feed these into the policy.
$K$ is set to the half of map and agent elements respectively.
Regarding the routes that capture lateral behaviors, we filter out the segments where the point closest to the current pose of the ego vehicle is the starting point, retaining $K_r$ points. In this case, $K_r$ is set to a quarter of $N_r$ points in one route. Finally, we transform the routes, agent, and map poses into the coordinate frame of the ego vehicle at the current time step $t$. We subtract the historical time steps $t-H:t$ from the current time step $t$, yielding time steps in range $-H:0$.

\noindent\textbf{Query-based Transformer decoder.} We employ the same backbone network architecture as the mode selector, but with different query dimensions. Due to the IVM and the fact that different modes yield distinct states, the map and agent information cannot be shared among modes. As a result, we fuse information for each individual mode. Specifically, the query dimension is \(1 \times D\), while the dimensions of the keys and values are \((N + N_{m}) \times D\). The output feature dimension remains \(1 \times D\).
Note that Transformer decoder can process information from multiple modes in parallel, eliminating the need to handle each mode sequentially.

\noindent\textbf{Policy output.} The mode feature is processed by two distinct heads: a policy head and a value head. Each head comprises its own MLP to produce the parameters for the action distribution and the corresponding value estimate. We employ a Gaussian distribution to model the action distribution, where actions are sampled from this distribution during training. In contrast, during inference, we utilize the mean of the distribution to determine the actions.

\subsubsection{Rule-augmented Selector}
This module is only utilized during inference and takes as input the initial state \( s_0 \), the multi-modal ego-planned trajectories, and the predicted future trajectories of agents. It calculates driving-oriented metrics such as safety, progress, comfort. A comprehensive score is obtained by the weighted sum of rule-based scores and the mode scores provided by the mode selector. The ego-planned trajectory with the highest score is selected as the output of the planner.

\subsection{Training}
We first train the non-reactive transition model and freeze the weights during the training of the mode selector and trajectory generator. Instead of feeding all modes to the generator, we apply a winner-takes-all strategy, wherein a positive mode is assigned based on the ego ground-truth trajectory and serves as a condition for the trajectory generator.

\noindent\textbf{Mode assignment.}
For the lateral mode, we assign the route closest to the endpoint of ego ground-truth trajectory as the positive lateral mode.
For the longitudinal mode, we partition the longitudinal space into \(N_{\text{lon}}\) intervals and assign the interval containing the endpoint of the ground-truth trajectory as the positive longitudinal mode.

\noindent\textbf{Reward function.} To handle diverse scenarios, we use the negative displacement error (DE) between the ego future pose and the ground truth as a universal reward. We also introduce additional terms to improve trajectory quality: collision rate and drivable area compliance. If the future pose collides or falls outside the drivable area, the reward is set to -1; otherwise, it is 0.

\noindent\textbf{Mode dropout.}
In some cases, there are no available routes for ego to follow. However, since routes serve as queries in Transformer, the absence of a route can lead to unstable or hazardous outputs. To mitigate this issue, we implement a mode dropout module during training that randomly masks routes to prevent over-reliance on this information.
% To prevent over-reliance on mode or route information due to Transformers' residual connections, we implement a mode dropout module during training that randomly masks the route to mitigate this issue.

\noindent\textbf{Loss function.}
For the selector, we use cross-entropy loss that is the negative log-likelihood of the positive mode and a side task that regresses the ego ground-truth trajectory.
For the generator, we use PPO~\cite{schulman2017proximal} loss that consists of three parts: policy improvement, value estimation, and entropy. Full description can be found in supplementary.

%% file: sections/4-results.tex
\section{Experiments}

\subsection{Experimental Setup}

\noindent\textbf{Dataset and simulator.} We use nuPlan~\cite{caesar2021nuplan}, a large-scale closed-loop platform for studying trajectory planning in autonomous driving, to evaluate the efficacy of our method. The nuPlan dataset contains driving log data over 1,500 hours collected by human expert drivers across 4 diverse cities. It includes complex, diverse scenarios such as lane follow and change, left and right turn, traversing intersections and bus stops, roundabouts, interaction with pedestrians, etc. As a closed-loop platform, nuPlan provides a simulator that uses scenarios from the dataset as initialization. During the simulation, traffic agents are taken over by log-replay (non-reactive) or an IDM~\cite{treiber2000idm} policy (reactive). The ego vehicle is taken over by user-provided planners. The simulator lasts for 15 seconds and runs at 10 Hz. At each timestamp, the simulator queries the planner to plan a trajectory, which is tracked by an LQR controller to generate control commands to drive the ego vehicle.

\noindent\textbf{Benchmarks and metrics.} We use two benchmarks: Test14-Random and Reduced-Val14 for comparing with other methods and analyzing the design choices within our method. The Test14-Random provided by PlanTF~\cite{cheng2024rethinking} contains 261 scenarios. The Reduced-Val14 provided by PDM~\cite{Dauner2023CORL} contains 318 scenarios.

We use the closed-loop score (CLS) provided by the official nuPlan devkit\footnote{\url{https://github.com/motional/nuplan-devkit}} to assess the performance of all methods.
The CLS score comprehends different aspects such as safety (S-CR, S-TTC), drivable area compliance (S-Area), progress (S-PR), comfort, etc.
Based on the different behavior types of traffic agents, CLS is detailed into CLS-NR (non-reactive) and CLS-R (reactive). 
% To further analyze various components of our methods, we also use open-loop metrics, such as loss of trajectory generator and selector.

\input{table/sota.tex}

\noindent\textbf{Implementation details.} We follow PDM~\cite{Dauner2023CORL} to construct our training and validation splits. The size of the training set is 176,218 where all available scenario types are used, with a number of 4,000 scenarios per type. The size of the validation set is 1,118 where 100 scenarios with 14 types are selected. We train all models with 50 epochs in 2 NVIDIA 3090 GPUs. The batch size is 64 per GPU. We use AdamW optimizer with an initial learning rate of 1e-4 and reduce the learning rate when the validation loss stops decreasing with a patience of 0 and decrease factor of 0.3. For RL training, we set the discount $\gamma = 0.1$ and the GAE parameter $\lambda = 0.9$. The weights of value, policy, and entropy loss are set to 3, 100, and 0.001, respectively. The number of longitudinal modes is set to 12 and a maximum number of lateral modes are set to 5.

\subsection{Comparison with SOTAs}

\input{table/ablation_rl.tex}

\noindent\textbf{SOTAs.} We categorize the methods into Rule, IL, and RL based on the type of trajectory generator.
(1) PDM~\cite{Dauner2023CORL} wins the nuPlan challenge 2023, its IL-based and rule-based variants are denoted as PDM-Open and PDM-Closed, respectively. PDM-Closed follows the generation-selection framework where IDM is used to generate multiple candidate trajectories and rule-based selector considering safety, progress, and comfort is used to select the best trajectory.
(2) PLUTO~\cite{cheng2024pluto} also obeys the generation-selection framework and uses contrastive IL to incorporate various data augmentation techniques and trains the generator.
(3) Gen-Drive~\cite{huang2024gen} is a concurrent work that follows a pretrain-finetune pipeline where IL is used to pretrain a diffusion-based planner and RL is used to finetune the denoising process based on a reward model trained by AI preference.

\noindent\textbf{Results.} We compare our method with SOTAs in Test14-Random and Reduced-Val14 benchmark as shown in \cref{table:main-results1} and \cref{table:main-results2}.
Overall, our CarPlanner demonstrates superior performance, particularly in non-reactive environments.

In the non-reactive setting, our method achieves the highest scores across all metrics, with an improvement of 4.02 and 2.15 compared to PDM-Closed and PLUTO, establishing the potential of RL and the superior performance of our proposed framework.
Moreover, CarPlanner reveals substantial improvement in the progress metric S-PR compared to PDM-Closed in \cref{table:main-results2} and comparable collision metric S-CR, indicating the ability of our method to improving driving efficiency while maintaining safe driving.
Importantly, we do not apply any techniques commonly used in IL such as data augmentation~\cite{cheng2024rethinking,cheng2024pluto} and ego-history masking~\cite{Guo-RSS-23}, underscoring the intrinsic capability of our approach to solving the closed-loop task.

In the reactive setting, while our method performs well, it falls slightly short of PDM-Closed. This discrepancy arises because our model was trained exclusively in non-reactive settings and has not interacted with the IDM policy used by reactive settings; as a result, our model is less robust to disturbances generated by reactive agents during testing.

\subsection{Ablation Studies}

\input{table/ablation_il_rl.tex}

We investigate the effects of different design choices in RL training. The results are shown in \cref{table:abla-rl}.

\noindent\textbf{Influence of reward items.}
% The results demonstrate that the DE and quality rewards are complementary.
When using the quality reward only, the planner tends to generate static trajectories and achieves a low progress metric. This occurs because the ego vehicle begins in a safe, drivable state, but moving forward is at risk of collisions or leaving the drivable area.
% On the other hand, compared to using DE reward only, incorporating the quality reward significantly improves closed-loop metrics.
On the other hand, when the quality reward is incorporated alongside the DE reward, it leads to significant improvements in closed-loop metrics compared to using the DE reward alone.
For instance, the S-CR metric rises from 97.49 to 99.22, and the S-Area metric rises from 96.91 to 99.22. These improvements indicate that the quality reward encourages safe and comfortable behaviors.

\noindent\textbf{Effectiveness of IVM.}
The results show that the coordinate transformation and KNN techniques in IVM notably improve closed-loop metrics and generator loss.
For instance, with the coordinate transformation technique, the overall closed-loop score increases from 90.78 to 94.07, and S-PR rises from 91.37 to 95.06.
These improvements are attributed to the enhanced accuracy of value estimation in RL, leading to generalized driving in closed-loop.

\begin{figure*}[ht]
    \begin{center}
    \includegraphics[width=0.98\textwidth]{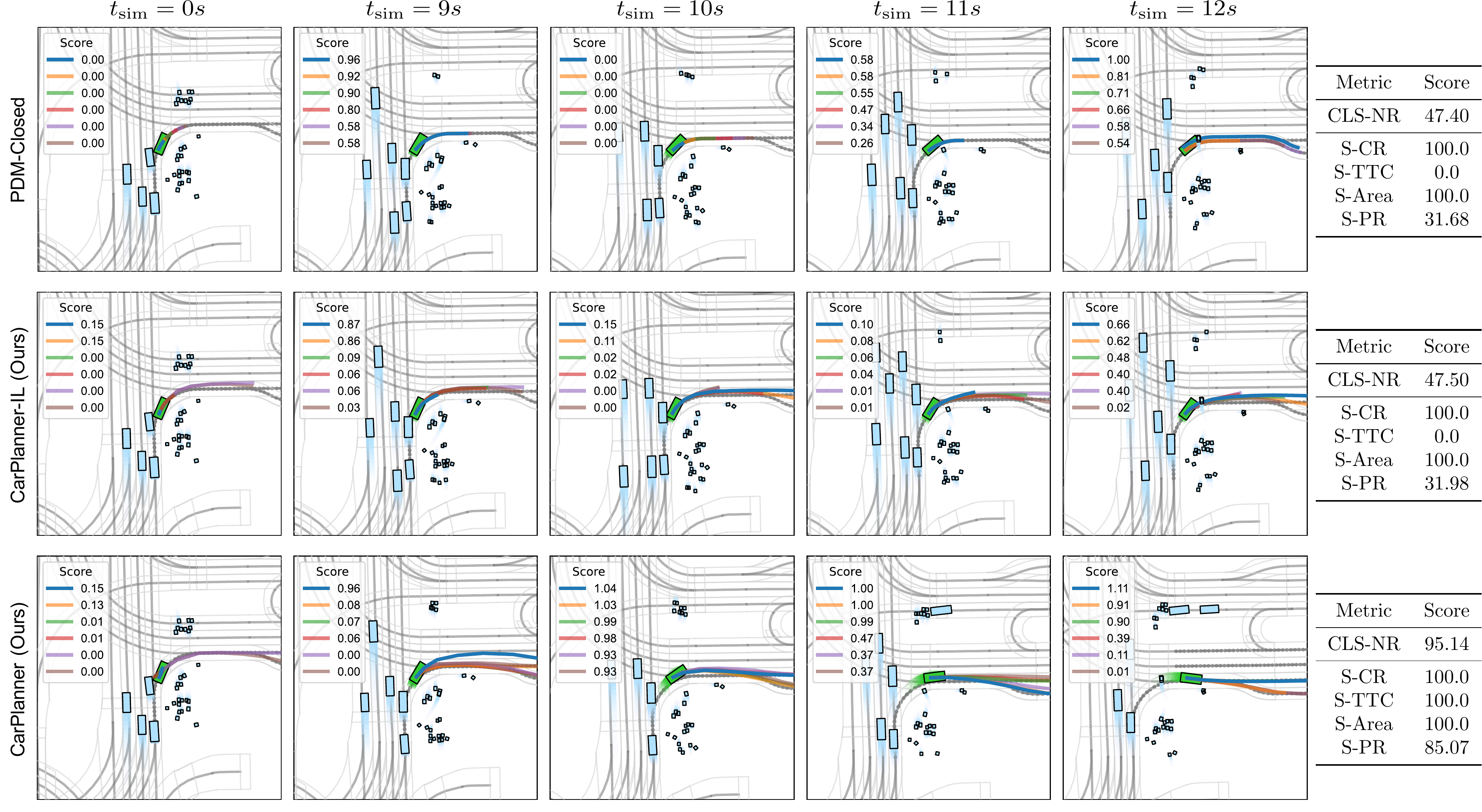}
    %%% trim={<left> <lower> <right> <upper>}
    \end{center}
    \vspace{-0.7cm}
    \caption{Qualitative comparison of PDM-Closed and our method in non-reactive environments. The scenario is annotated as \texttt{waiting\_for\_pedestrian\_to\_cross}. In each frame shot, ego vehicle is marked as {\color{LimeGreen} green}. Traffic agents are marked as {\color{DeepSkyBlue} sky blue}. Lineplot with {\color{Blue} blue} is the ego planned trajectory.
    }
    \vspace{-0.5cm}
    \label{figure:vis}
\end{figure*}

\subsection{Extention to IL}
\label{section:abla-il}

In addition to designing for RL training, we also extend the CarPlanner to incorporate IL. We conduct rigorous analysis to compare the effects of various design choices in IL and RL training, as summarized in \cref{table:abla-il-rl}. Our findings indicate that while mode dropout and selector side task contribute to both IL and RL training, ego-history dropout and backbone sharing, often effective in IL, are less suitable for RL.

\noindent\textbf{Ego-history dropout.} Previous works~\cite{Ogale-RSS-19, Guo-RSS-23, cheng2024rethinking, cheng2024pluto} suggest that planners trained via IL may rely too heavily on past poses and neglect environmental state information. To counter this, we combine techniques from ChauffeurNet~\cite{Ogale-RSS-19} and PlanTF~\cite{cheng2024rethinking} into an ego-history dropout module, randomly masking ego history poses and current velocity to alleviate the causal confusion issue.

Our experiments confirm that ego-history dropout benifits IL training, as it improves performance across closed-loop metrics like S-CR and S-Area. However, in RL training, we observe a negative impact on advantage estimation due to ego-history dropout, which significantly affects the value part of generator loss, leading to closed-loop performance degradation.
This suggests that RL training naturally addresses the causal confusion problem inherent in IL by uncovering causal relationships that align with the reward signal, which explicitly encodes task-oriented preferences. This capability highlights the potential of RL to push the boundaries of learning-based planning.

\noindent\textbf{Backbone sharing.} This choice, often used in IL-based multi-modal planners, promotes feature sharing across tasks to improve generalization. While backbone sharing helps IL by balancing losses across trajectory generator and selector, we find it adversely affects RL training. Specifically, backbone sharing leads to higher losses for both the trajectory generator and selector in RL, indicating that gradients from each task interfere. The divergent objectives in RL for trajectory generation and selection tasks seem to conflict, reducing overall policy performance. Consequently, we avoid backbone sharing in our RL framework to maintain task-specific gradient flow and improve policy quality.

\subsection{Qualitative Results}

We provide qualitative results as shown in \cref{figure:vis}. In this scenario, ego vehicle is required to execute a right turn while navigating around pedestrians. In this case, Our method shows a smooth, efficient performance.
From $t_{\text{sim}} = 0s$ to $t_{\text{sim}} = 9s$, all methods wait for the pedestrians to cross the road.
At $t_{\text{sim}} = 10s$, an unexpected pedestrian goes back and prepares to re-cross the road. PDM-Closed is unaware of this situation and takes an emergency stop, but it still intersects with this pedestrian. In contrast, our IL variant displays an awareness of the pedestrian's movements and consequently conducts a braking maneuver. However, it still remains close to the pedestrian. Our RL method avoids this hazard by starting up early up to $t_{\text{sim}} = 9s$ and achieves the highest progress and safety metrics.

%% file: table/sota.tex
\begin{table}[t]
    \centering
    
    \begin{minipage}{0.48\textwidth}
      \centering
      \setlength{\tabcolsep}{7pt}
      \setlength{\aboverulesep}{0.6pt}
      \setlength{\belowrulesep}{0.6pt}
      \fontsize{5pt}{5pt}\selectfont
      \resizebox{\textwidth}{!}{\begin{tabular}{c | l | c c}
        \toprule
        Type & {Planners} & CLS-NR & {CLS-R} \\
        \midrule
        \multirow{2}{*}{\makecell{Rule}} & {IDM~\cite{treiber2000idm}} & 70.39 & {72.42} \\
        & {PDM-Closed~\cite{Dauner2023CORL}} & {90.05} & \textbf{91.64} \\
        \midrule
        \multirow{8}{*}{\makecell{IL}} & {RasterModel~\cite{caesar2021nuplan}} & 69.66 & {67.54} \\
        & {UrbanDriver~\cite{scheel2022urban}} & 63.27 & {61.02} \\
        & {GC-PGP~\cite{hallgarten2023prediction}} & 55.99 & {51.39} \\
        & {PDM-Open~\cite{Dauner2023CORL}} & 52.80 & {57.23} \\
        & {GameFormer~\cite{huang2023gameformer}} & 80.80 & {79.31} \\
        & {PlanTF~\cite{cheng2024rethinking}} & 86.48 & {80.59} \\
        & {PEP~\cite{zhang2024pep}} & {91.45} & {89.74} \\
        & {PLUTO~\cite{cheng2024pluto}} & \underline{91.92} & 90.03 \\
        \midrule
        \multirow{1}{*}{\makecell{RL}} & {CarPlanner (Ours)} & \textbf{94.07} & \underline{91.1} \\
        \bottomrule
      \end{tabular}}
    \vspace{-0.3cm}
    \caption{Comparison with SOTAs in Test14-Random. Based on the type of trajectory generator, all methods are categorized into Rule, IL, and RL. The best result is in \textbf{bold} and the second best result is \underline{underlined}.}
    \vspace{-0.2cm}
    \label{table:main-results1}
    \end{minipage}

    \hfill
    \vspace{0.3cm}

    \begin{minipage}{0.48\textwidth}
      \centering
      \setlength{\tabcolsep}{4pt}
      \setlength{\aboverulesep}{0.5pt}
      \setlength{\belowrulesep}{0.5pt}
      \fontsize{5pt}{6pt}\selectfont
      \resizebox{\textwidth}{!}{\begin{tabular}{c | l | c c c }
        \toprule
        Type & {Planners} & CLS-NR & {S-CR} & S-PR \\
        \midrule
        \multirow{1}{*}{\makecell{Rule}} & {PDM-Closed~\cite{Dauner2023CORL}} & \underline{91.21} & \textbf{97.01} & \underline{92.68} \\
        \midrule
        \multirow{3}{*}{\makecell{IL}} & {GameFormer~\cite{huang2023gameformer}} & 83.76 & {94.73} & 88.12 \\
        & {PlanTF~\cite{cheng2024rethinking}} & 83.66 & {94.02} & 92.67 \\
        & {Gen-Drive (Pretrain)~\cite{huang2024gen}} & 85.12 & 93.65 & 86.64 \\
        % & {CarPlanner-IL (Ours)} & 90.34 & 95.11 &  \underline{94.57} \\
        \midrule
        \multirow{2}{*}{\makecell{RL}} & {Gen-Drive (Finetune)~\cite{huang2024gen}} & 87.53 & 95.72 & 89.94 \\
        & {CarPlanner (Ours)} & \textbf{91.45} & \underline{96.38} & \textbf{95.37} \\
        \bottomrule
      \end{tabular}}
        \vspace{-0.3cm}
      \caption{Comparison with SOTAs in Reduced-Val14 with non-reactive traffic agents.}
        \vspace{-0.5cm}
      \label{table:main-results2}
    \end{minipage}
\end{table}

%% file: table/ablation_rl.tex
\begin{table*}[t]
  % \vspace{6pt}
  \centering
  % \vspace{-0.2cm}
  \setlength{\tabcolsep}{4pt} % Adjust the space between columns
  \setlength{\aboverulesep}{0.3pt}
  \setlength{\belowrulesep}{0.3pt}
  % \small
  \fontsize{5pt}{6pt}\selectfont % Set font size and line spacing
  \resizebox{0.98\textwidth}{!}{\begin{tabular}{ c c c c |  c c c c c | c c }
    \toprule
  \multicolumn{4}{c|}{Design Choices} & \multicolumn{5}{c|}{Closed-loop metrics ($\uparrow$)}  &  \multicolumn{2}{c}{Open-loop losses ($\downarrow$)}  \\
  \midrule
  \makecell{Reward\\DE} &  \makecell{Reward\\Quality} & \makecell{Coord\\Trans} & \makecell{KNN} &  CLS-NR &  S-CR & S-Area & S-PR & S-Comfort  & \makecell{Loss\\Selector} & \makecell{Loss\\Generator}  \\
  \midrule
  % \multirow{2}{*}{\makecell{RL}}
      {\xmark}   & {\cmark}   & {\cmark}    & {\cmark}  &  31.79                & 95.74              & 98.45               & 33.10              & 48.84               &  1.03              &  \textbf{30.3}  \\ 
      {\cmark}   & {\xmark}   & {\cmark}    & {\cmark}  &  90.44                & 97.49              & 96.91               & 93.33              & 90.73               &  \textbf{0.99}     &  \underline{1221.6}  \\ 
      {\cmark}   & {\cmark}   & {\xmark}    & {\cmark}  &  90.78                & 96.92              & \underline{98.46}   & 91.37              & \textbf{94.23}      &  \underline{1.00}  &  2130.7  \\ 
      {\cmark}   & {\cmark}   & {\cmark}    & {\xmark}  &  \underline{92.73}    & \underline{98.07}  & \underline{98.46}   & \underline{94.69}  & \underline{93.44}   &  1.03              &  2083.6  \\ 
      \rowcolor{gray!30} {\cmark}   & {\cmark}   & {\cmark}    & {\cmark}  &  \textbf{94.07}       & \textbf{99.22}     & \textbf{99.22}      & \textbf{95.06}     & 91.09               &  1.03              &  1624.5  \\ 
      \bottomrule
  \end{tabular}}
  \vspace{-0.3cm}
  \caption{Ablation studies on the design choices in RL training. Results are in Test14-random non-reactive benchmark.}
  \label{table:abla-rl}
  \vspace{-0.6cm}
\end{table*}

%% file: table/ablation_il_rl.tex
\begin{table*}[t]
    % \vspace{6pt}
    \centering
    % \vspace{-0.2cm}
    \setlength{\tabcolsep}{3pt} % Adjust the space between columns
    \setlength{\aboverulesep}{0.3pt}
    \setlength{\belowrulesep}{0.3pt}
    % \small
    \fontsize{5pt}{6pt}\selectfont % Set font size and line spacing
    \resizebox{0.98\textwidth}{!}{\begin{tabular}{c | c c c c |  c c c c c | c c }
    \toprule
    & \multicolumn{4}{c|}{Design Choices} & \multicolumn{5}{c|}{Closed-loop metrics ($\uparrow$)}  &  \multicolumn{2}{c}{Open-loop losses ($\downarrow$)}  \\
    \midrule
    \makecell{Loss\\Type} & \makecell{Mode\\Dropout}   &  \makecell{Selector\\Side Task}  &  \makecell{Ego-history\\Dropout} & \makecell{Backbone\\Sharing}   &  CLS-NR &  S-CR  & S-Area  & S-PR & S-Comfort   & \makecell{Loss\\Selector} & \makecell{Loss\\Generator}  \\
    \midrule
    \multirow{5}{*}{\makecell{IL}} 
        & {\xmark}   & {\xmark}   & {\xmark}    & {\xmark}    &  90.82             & 97.29              & 98.45                & 92.15                & 94.57                 &  \underline{1.04}    &  \textbf{147.5}    \\ 
        & {\cmark}   & {\xmark}   & {\xmark}    & {\xmark}    &  91.21             & 96.54              & 98.46                & 91.44                & \textbf{96.92}        &  1.07                &  \underline{153.0}    \\ 
        & {\cmark}    & {\cmark}  & {\xmark}   & {\xmark}     &  91.51             & 96.91              & 98.46                & \textbf{95.30}       & \underline{96.91}     &  \underline{1.04}    &  162.3    \\ 
        & {\cmark}    & {\cmark}  & {\cmark}   & {\xmark}     &  92.72             & 98.06              & 98.84                & 94.88                & 95.35                 &  \underline{1.04}    &  167.5    \\ 
        \rowcolor{gray!30} & {\cmark}    & {\cmark}  & {\cmark}   & {\cmark}     &  93.41             & \underline{98.85}  & 98.85                & 93.87                & 96.15                 &  \underline{1.04}    &  174.3    \\ 
    \midrule
    % \multirow{5}{*}{\makecell{RL}}
        & {\xmark}    & {\xmark}    & {\xmark}   & {\xmark}   &  91.67             & 98.84              & 98.84                & 91.69                & 90.73                 &  \underline{1.04}    &  1812.6  \\ 
        & {\cmark}    & {\xmark}    & {\xmark}   & {\xmark}   &  \underline{93.46} & 98.07              & \textbf{99.61}       & 94.26                & 92.28                 &  1.09                &  2254.6  \\ 
        \rowcolor{gray!30}
        & {\cmark}    & {\cmark}    & {\xmark}   & {\xmark}   &  \textbf{94.07}    & \textbf{99.22}     & \underline{99.22}    & \underline{95.06}    & 91.09                 &  \textbf{1.03}       &  1624.5  \\ 
        \multirow{-3}{*}{\makecell{RL}}
        & {\cmark}    & {\cmark}    & {\cmark}   & {\xmark}   &  89.51             & 97.27              & 98.44                & 90.93                & 83.20                 &  1.05                &  5424.3  \\ 
        & {\cmark}    & {\cmark}    & {\cmark}   & {\cmark}   &  88.66             & 95.54              & 98.84                & 92.82                & 86.05                 &  1.21                &  1928.1  \\ 
        \bottomrule              
    \end{tabular}}
  \vspace{-0.3cm}
    \caption{Effect of different components on IL and RL loss using our CarPlanner. Results are in Test14-random non-reactive benchmark.}
    \label{table:abla-il-rl}
    \vspace{-0.3cm}
\end{table*}

%% file: sections/5-conclusion.tex
\section{Conclusion}

In this paper, we introduce CarPlanner, a consistent auto-regressive planner aiming at large-scale RL training.   
Thanks to the proposed framework, we train an RL-based planner that outperforms existing RL-, IL-, and rule-based SOTAs.
Furthermore, we provide analysis indicating the characteristics of IL and RL, highlighting the potential of RL to take a further step toward learning-based planning.

\noindent\textbf{Limitations and future work.}
RL needs delicate design and is prone to input representation.
RL can overfit its training environment and suffer from performance drop in unseen environments~\cite{kirk2023survey}. 
Our method leverages expert-aided reward design to guide exploration. However, this approach may constrain the full potential of RL, as it inherently relies on expert demonstrations and may hinder the discovery of solutions that surpass human expertise.
Future work aims to develop robust RL algorithms capable of overcoming these limitations, enabling autonomous exploration and generalization across diverse environments.

\section{Acknowledgement}

Many thanks to Jingke Wang for helpful discussions and all reviewers for improving the paper.
This work was supported by
Zhejiang Provincial Natural Science Foundation of China under Grant No. LD24F030001, and by
the National Nature Science Foundation of China under Grant 62373322.

%% file: sections/6-supple.tex
\clearpage
\maketitlesupplementary
\appendix

\section{Training Procedure}

\cref{alg:training} outlines the training process for the CarPlanner framework.
% The procedure involves two primary steps: 1)~training the non-reactive transition model, and 2)~training the mode selector and the trajectory generator.
Notably, during the training of the trajectory generator, we have the flexibility to employ either RL or IL, but, in this work, we do not combine RL and IL simultaneously, opting instead to explore their distinct characteristics separately.
The definitions of the loss functions are given in the following.

\noindent\textbf{Loss of non-reactive transition model.}
The non-reactive transition model $\beta$ is trained to simulate agent trajectories based on the initial state $\boldsymbol{s}_0$. For each data sample $(\boldsymbol{s}_0, s^{1:N,\text{gt}}_{1:T}) \in \mathcal{D}$, the model predicts trajectories $s^{1:N}_{1:T} = \beta(\boldsymbol{s}_0)$, and the training objective minimizes the L1 loss:
\begin{equation}
L_{\text{tm}} = \frac{1}{T} \sum_{t=1}^T \sum_{n=1}^N \left\| s_t^n - s_t^{n,\text{gt}} \right\|_1.
\end{equation}

\noindent\textbf{Mode selector loss.}
This contains two parts: cross-entropy and side-task loss.
The cross-entropy loss is defined as:
\begin{equation}
    \text{CrossEntropyLoss}(\boldsymbol{\sigma}, c^*) = -\sum_{i=1}^{N_{\text{mode}}} \mathbb{I}(c_i = c^*) \log \sigma_i,
\end{equation}
where $\sigma_i$ is the assigned score for mode $c_i$, $N_{\text{mode}}$ is the number of candidate modes, and $\mathbb{I}$ is the indicator function.
The side-task loss is defined as:
\begin{equation}
    \text{SideTaskLoss}(\bar{s}^{0}_{1:T}, s^{0,\text{gt}}_{1:T}) = \frac{1}{T} \sum_{t=1}^T \left\| \bar{s}_t^0 - s_t^{0,\text{gt}} \right\|_1,
\end{equation}
where $\bar{s}_t^0$ is the output ego future trajectory.

\noindent\textbf{Generator loss with RL.}
The PPO~\cite{schulman2017proximal} loss consists of three parts: policy, value, and entropy loss.
The policy loss is defined as:
\begin{equation}
\begin{split}
    &\text{PolicyLoss}(a_{0:T-1}, d_{0:T-1, \text{new}}, d_{0:T-1}, A_{0:T-1}) \\
    = &- \frac{1}{T} \sum_{t=0}^{T-1} \min \left( r_t A_t, \text{clip}(r_t, 1-\epsilon, 1+\epsilon) A_t \right),
\end{split}
\end{equation}

\input{blocks/algorithm.tex}

where the ratio $r_t$ is given by $r_t = \frac{\text{Prob}(a_t, d_{t,\text{new}})}{\text{Prob}(a_t, d_{t})}$, $d_{t,\text{new}}$ and $d_t$ are the policy distributions (mean and standard deviation of Gaussian distribution) at time step $t$ induced by $\pi$ and $\pi_{\text{old}}$ respectively, the function $\text{Prob}(a, d)$ calculates the probability of a given action $a$ under a distribution $d$, and $A_t$ is the advantage estimated using GAE~\cite{schulman2015high}.
The value and entropy loss are defined as:
\begin{equation}
    \text{ValueLoss}(V_{0:T-1, \text{new}}, \hat{R}_{0:T-1}) = \frac{1}{T} \sum_{t=0}^{T-1} \left\| V_{t,\text{new}} - \hat{R}_t \right\|_2^2,
\end{equation}
\begin{equation}
    \text{Entropy}(d_{0:T-1, \text{new}}) = \frac{1}{T} \sum_{t=0}^{T-1} \mathcal{H}(d_{t,\text{new}}),
\end{equation}
\noindent where $V_{t,\text{new}}$ and $\hat{R}_t$ are the predicted and actual returns, and $\mathcal{H}$ represents the entropy of the policy distribution $d$.

\noindent\textbf{Generator loss with IL.}
In IL, the generator minimizes the trajectory error between the ego-planned trajectory $s^0_{1:T}$ and the ground-truth trajectory $s^{0,\text{gt}}_{1:T}$. The loss is defined as:
\begin{equation}
    L_{\text{generator}} = \frac{1}{T} \sum_{t=1}^T \left\| s_t^0 - s_t^{0,\text{gt}} \right\|_1.
\end{equation}

\section{Implementation Details}

The hyperparameters of model architecture, PPO-related parameters, and loss weights are summarized in \cref{table:param}. The magnitudes of value, policy, and entropy loss are $10^3$, $10^0$, and $10^{-3}$, respectively. The trajectory generator generates trajectories with a time horizon of 8 seconds at 1-second intervals, corresponding to time horizon $T = 8$. During testing, these trajectories are interpolated to 0.1-second intervals.
The weight of scores generated by the rule and mode selectors is set to a ratio of $1:0.3$.
In cases where no ego candidate trajectory satisfies the safety criteria evaluated by the rule selector, an emergency stop is triggered.
For the Test14-Random benchmark, a replanning frequency of 10Hz is employed, adhering to the official nuPlan simulation configuration. In contrast, for the Reduced-Val14 benchmark, a replanning frequency of 1Hz is used to ensure a fair comparison with Gen-Drive~\cite{huang2024gen}.

\begin{table}[t]
\centering
% \captionsetup{font=large}
\centering
    \begin{tabular}{l c}
    \toprule
    {Parameter} & Value \\
    \midrule
    Feature dimension $D$ & 256 \\
    Static point dimension $D_m$ & 9 \\
    Agent pose dimension $D_a$ & 10 \\
    Activation & ReLU \\
    Number of layers & $3$ \\
    Number of attention heads & 8 \\
    % Hidden size & 512 \\
    % Dimension of key and value & 32 \\
    Dropout & 0.1 \\
    \midrule
    discount factor $\gamma$ & 0.1 \\
    GAE parameter $\lambda$ & 0.9 \\
    Clip range $\epsilon$ & 0.2 \\
    Update interval $I$ & 8 \\
    \midrule
    Weight of selector loss & 1 \\
    Weight of value loss & 3 \\
    Weight of policy loss & 100 \\
    Weight of entropy loss & 0.001 \\
    Weight of IL loss & 1 \\
    \bottomrule
    \end{tabular}
    \caption{Hyperparameters of model architecture, PPO-related parameters, and loss weights.}
    \label{table:param}
\end{table}%

\input{blocks/table_supple_rl.tex}

\section{Ablation Study on RL Training}

\input{blocks/figure_time_horizon.tex}

\input{blocks/figure_vis2.tex}

In this part, we examine the training efficiency of CarPlanner, performance of vanilla and consistent auto-regressive frameworks, the use of reactive and non-reactive model in RL training, and the impact of varying the time horizon.

\noindent\textbf{Training efficiency.}
We compare the efficiency of our model-based framework with that of ScenarioNet~\cite{li2023scenarionet}, which is an open-source platform for model-free RL training in real-world datasets~\cite{ettinger2021large, caesar2021nuplan}.
As shown in \cref{table:efficiency}, CarPlanner achieves a remarkable improvement in sampling efficiency, outperforming ScenarioNet by two orders of magnitude. Furthermore, CarPlanner not only excels in efficiency but also achieves SOTA performance, surpassing ScenarioNet by a wide margin.
% This indicates that our framework achieves greater training efficiency.

\noindent\textbf{Vanilla vs. consistent auto-regressive framework.}
The results are shown in \cref{table:supple-abla-rl-reward,table:consistency}.
The consistent auto-regressive framework generates multi-modal trajectories by conditioning on mode representations. In contrast, the vanilla framework relies on random sampling from the action Gaussian distribution to produce multi-modal trajectories. To ensure comparability in the number of modes generated by both frameworks, we sample 60 trajectories in parallel for the vanilla framework.
Given that random sampling introduces variability, we average the results across 3 random seeds.
For the consistent framework, we use displacement error (DE) and final error (FE) as guide functions to assist the policy in generating mode-aligned trajectories.
For the vanilla framework, DE is compared against a progress reward, which encourages longitudinal movement along the route while discouraging excessive lateral deviations that move the vehicle too far from any possible route.
The consistent ratio computes the ratio of generated trajectories that fall in their corresponding modes in longitudinal and lateral directions separately.

Overall, our proposed consistent framework outperforms the vanilla framework in terms of closed-loop performance, highlighting the benefits of incorporating consistency. Furthermore, RL provides more consistant trajectories than the vanilla framework and IL-based methods. Additionally, we find that DE serves as an effective guide function for policy training, further enhancing closed-loop performance.

\noindent\textbf{Reactive vs. non-reactive transition model.}
We compare the performance of the CarPlanner framework when trained with reactive and non-reactive transition models. The reactive transition model shares a similar architecture with the auto-regressive planner for the ego vehicle, utilizing relative pose encoding~\cite{zhang2024real} as the backbone network to extract features of traffic agents and predict their subsequent poses. The training loss and hyperparameters are consistent with those used for the non-reactive transition model.
As shown in \cref{table:supple-abla-rl-tm}, except for the S-Area metric, using non-reactive transition model outperforms the reactive transition model in our current implementation.
The primary difference lies in the assumptions about traffic agents: the reactive transition model assumes that the ego vehicle can negotiate with traffic agents and share the same priority, whereas in the non-reactive model, traffic agents do not respond to the ego vehicle, effectively assigning them higher priority.
A representative example is presented in \cref{figure:vis2}.
When trained with the reactive transition model, the planner assumes pedestrians will yield to the vehicle, leading it to attempt to move forward.
However, at $t_{\text{sim}} = 12s$, the planner collides with pedestrians, triggering an emergency brake, which negatively impacts safety, progress, and comfort metrics.
Although the performance of using reactive transition model is not satisfied currently, it is a more realistic assumption and we will further investigate this in future work.

\noindent\textbf{Time horizon.} We evaluate the CarPlanner framework by training it with different time horizons, including 1, 3, 5, and 8 seconds, and testing the planners in each time horizon.
The results in \cref{figure:supple-time-horizon} confirm that increasing the time horizon has a positive effect on the performance for both training and testing.
A special case is when the training time horizon is set to 1, all tested time horizons exhibit poor performance, highlighting the importance of multi-step learning in RL.
Additionally, the observation that increasing the training time horizon enhances closed-loop performance suggests the potential for further improvements by extending the time horizon beyond 8 seconds.
However, due to current limitations in data preparation, which is designed for horizons up to 8 seconds, expanding the time horizon would not provide map information or ground-truth trajectories, hindering further analysis. Consequently, we leave this exploration for future work.

\section{Comparison with Differentiable Loss}

\input{blocks/table_supple_diff_loss.tex}

\input{blocks/figure_diff_loss.tex}

In typical IL setting, the supervision signal provided to the trajectory generator is the displacement error (DE) between the ego-planned trajectory and the ground-truth trajectory. Several works~\cite{suo2021trafficsim, huang2023gameformer, cheng2024pluto} propose to convert non-differentiable metrics, such as avoiding collision (Col) and adherence to drivable area (Area), into differentiable loss functions that can directly backpropagate to the generator. In contrast, CarPlanner leverages an RL framework, which introduces surrogate objectives to indirectly optimize these non-differentiable metrics.

In this part, we compare these two approaches which provide rich supervision signals to the trajectory generator. The results are summarized in \cref{table:supple-abla-diff-loss}.
In IL training, the Col and Area metrics are converted into differentiable loss functions, whereas in RL training, Col and Area are treated as reward functions, contributing to the quality reward as described in the main paper.
It is important to note that the implementations for differentiable loss functions and reward functions are identical, except that gradient flow is enabled for differentiable loss functions.
The open-loop metrics compute the Col and Area values across all candidate multi-modal trajectories, with the Mean, Min, and Max referring to the mean, minimum, and maximum values of the Col and Area metrics within the candidate trajectory set.

Our findings suggest that incorporating Col loss benefits the open-loop Col metric and improves the closed-loop S-CR metrics, thereby enhancing closed-loop performance. However, incorporating Area loss results in better open-loop Area metrics but deteriorates closed-loop performance. Compared to differentiable loss functions, RL with Col and Area as quality rewards yields the trajectory set with the highest overall quality, as evidenced by smaller Mean and Max metrics in open-loop metrics.
This improvement can be attributed to RL's ability to optimize the reward-to-go using surrogate objectives that account for future rewards, while differentiable loss functions are limited to timewise-aligned optimization in our current implementation.
This distinction is illustrated in \cref{figure:diff-loss}: in (a), the loss at time step $t$ is directly computed from $s^0_t$, meaning that during backward propagation, the loss at time step $t$ cannot influence the optimization of prior time steps.
In (b), however, the non-differentiable reward is aggregated into a return (reward-to-go), which serves as a reference for computing the loss at time step $t$. Through this process, the reward at time step $t$ can influence the trajectory at earlier time steps $t'$ ($t' < t$). In the future, we aim to combine the advantages of differentiable loss which can provide low-variance gradients, and RL which can provide long-term foresight, by model-based RL optimization techniques~\cite{claveramodel, hansen2022temporal}.

\section{Effect of Mode Representation}

In this part, we examine the impact of mode representations on performance. The results are presented in \cref{table:supple-abla-mode}.
For both the vanilla and consistent frameworks, we disable the use of random sampling to focus solely on mode-aligned trajectories. As a result, the vanilla framework can only generate single-modal trajectories, leading to the lowest performance.
In the consistent framework, we explore two types of mode representations: Lon and Lon-Lat. The Lon representation assigns modes based on longitudinal movements along the route, whereas the Lon-Lat representation decomposes modes by both longitudinal and lateral movements.
Aligned with the main paper, we use ego-history dropout and backbone sharing only for IL training. For the Lon representation, we close mode dropout since it does not rely on any map or agent representation in initial state.
The results indicate that introducing consistency provides greater benefits to RL training, with the Lon-Lat representation proving to be more effective than the Lon representation. This suggests that decomposing mode representations into both longitudinal and lateral components enhances the model's ability by providing more explicit mode information.

% \section{Qualitative Results}

% Qualitative behaviors of our proposed CarPlanner can be found in \texttt{video.mp4}.

%% file: blocks/algorithm.tex
\newcommand{\rightcomment}[1]{\hfill\textcolor{gray}{\% #1}}
\newcommand{\comment}[1]{\NoNumber{\textcolor{gray}{\% #1}}}

\begin{algorithm}[H]
\begin{algorithmic}[1]
\caption{Training Procedure of CarPlanner}
\label{alg:training}

\small

\State \textbf{Input:} Dataset $\mathcal{D}$ containing initial states $\boldsymbol{s}_0$ and ground-truth trajectories $s^{0:N,\text{gt}}_{1:T}$, longitudinal modes $\boldsymbol{c}_{\text{lon}}$, discount factor $\gamma$, GAE parameter $\lambda$, update interval $I$.
\State \textbf{Require:} Non-reactive transition model $\beta$, mode selector $f_{\text{selector}}$, policy $\pi$, policy old $\pi_{\text{old}}$.

\State \textbf{\textcolor{gray}{Step 1: Training Transition Model}}
\For{$(\boldsymbol{s}_0, s^{1:N,\text{gt}}_{1:T}) \in \mathcal{D}$}
    \State Simulate agent trajectories $s^{1:N}_{1:T} \gets \beta(\boldsymbol{s}_0)$
    \State Calculate loss $L_{\text{tm}} \gets \text{L1Loss}(s^{1:N}_{1:T}, s^{1:N,\text{gt}}_{1:T})$
    \State Backpropagate and update $\beta$ using $L_{\text{tm}}$
\EndFor

\State \textbf{\textcolor{gray}{Step 2: Training Selector and Generator}}
\State Initialize {training\_step} $\gets 0$
\State Initialize policy old $\pi_{\text{old}} \gets \pi$
\For{$(\boldsymbol{s}_0, s^{0,\text{gt}}_{1:T}) \in \mathcal{D}$}
    \State \textbf{Non-Reactive Transition Model:} 
    \State Simulate agent trajectories $s^{1:N}_{1:T} \gets \beta(\boldsymbol{s}_0)$

    \State \textbf{Mode Assignment:}
    \State Determine $\boldsymbol{c}_{\text{lat}}$ based on $\boldsymbol{s}_0$
    \State Concatenate $\boldsymbol{c}_{\text{lat}}$ and $\boldsymbol{c}_{\text{lon}}$ to get $\boldsymbol{c}$
    \State Determine positive mode $c^*$ based on $s^{0,\text{gt}}_{1:T}$ and $\boldsymbol{c}$
    
    \State \textbf{Mode Selector Loss:}
    \State Compute scores $\boldsymbol{\sigma}, \bar{s}^{0}_{1:T} \gets f_{\text{selector}}(\boldsymbol{s}_0, \boldsymbol{c})$
    \State $L_{\text{selector}} \gets \text{CrossEntropyLoss}(\boldsymbol{\sigma}, c^*) + \text{SideTaskLoss}(\bar{s}^{0}_{1:T}, s^{0,\text{gt}}_{1:T})$

    \State \textbf{Generator Loss:}
    \If{Reinforcement Learning (RL) Training}
        \State Use $\pi_{\text{old}}$, $\boldsymbol{s}_0$, $c^*$, and $s^{1:N}_{1:T}$ to collect rollout data $(\boldsymbol{s}_{0:T-1}, a_{0:T-1}, d_{0:T-1}, V_{{0:T-1}}, R_{0:T-1})$
        \State Compute advantage $A_{0:T-1}$ and return $\hat{R}_{0:T-1}$ using GAE~\cite{schulman2015high}: $A_{0:T-1}, \hat{R}_{0:T-1} \gets \text{GAE}(R_{0:T-1}, V_{{0:T-1}}, \gamma, \lambda)$
        \State Compute policy distribution and value estimates: $(d_{0:T-1, \text{new}}, V_{0:T-1, \text{new}}) \gets \pi(\boldsymbol{s}_{0:T-1}, a_{0:T-1}, c^*)$
        \State $L_{\text{generator}} \gets \text{ValueLoss}(V_{0:T-1, \text{new}}, \hat{R}_{0:T-1}) + \text{PolicyLoss}(d_{0:T-1, \text{new}}, d_{0:T-1}, A_{0:T-1}) - \text{Entropy}(d_{0:T-1, \text{new}})$
    \ElsIf{Imitation Learning (IL) Training}
        \State Use $\pi$, $\boldsymbol{s}_0$, $c^*$, and $s^{1:N}_{1:T}$ to collect action sequence $a_{0:T-1}$
        \State Stack action sequence as ego-planned trajectory $s^{0}_{1:T} \gets \text{Stack}(a_{0:T-1})$
        \State $L_{\text{generator}} \gets \text{L1Loss}(s^{0}_{1:T}, s^{0,\text{gt}}_{1:T})$
    \EndIf
    
    \State \textbf{Overall Loss:}
    \State $L \gets L_{\text{selector}} + L_{\text{generator}}$
    \State Backpropagate and update $f_{\text{selector}}, \pi$ using $L$

    \State \textbf{Policy Update:}
    \State Increment {training\_step} $\gets$ {training\_step} $+ 1$
    \If{{training\_step} $\%$ $I$ $== 0$}
        \State Update $\pi_{\text{old}} \gets \pi$
    \EndIf
\EndFor

\end{algorithmic}
\end{algorithm}

%% file: blocks/table_supple_rl.tex
\begin{table}[t]
    % \vspace{6pt}
    % \centering
    % % \vspace{-0.2cm}
    \setlength{\tabcolsep}{2pt} % Adjust the space between columns
    % \setlength{\aboverulesep}{0.02pt}
    % \setlength{\belowrulesep}{0.02pt}
    % \small
    \fontsize{8pt}{10pt}\selectfont % Set font size and line spacing
    \resizebox{0.48\textwidth}{!}{\begin{tabular}{ c |  c | c c c }
    \toprule
    \makecell{Planner}   &  CLS-NR ($\uparrow$) &  \makecell{Efficiency\\(samples/sec, $\uparrow$)}  & Num. Samples & Train Time    \\
    \midrule
        {ScenarioNet~\cite{li2023scenarionet}}        & 55.60  & 25.72    & 7,798,472   & 3d12h11m38s  \\
        CarPlanner-IL     & \underline{93.41}  & \underline{1,181.46} & 70,487,200  & 16h34m12s    \\
        CarPlanner     & \textbf{94.07}  & \textbf{1,632.25} & 70,487,200  & 11h59m44s    \\
        \bottomrule
    \end{tabular}}
    % \vspace{-0.35cm}
    \caption{Comparison of training efficiency with model-free settings. Experimental results are based on the Test14-Random non-reactive benchmark.}
    \label{table:efficiency}
    % \vspace{-0.4cm}
\end{table}

\begin{table}[t]
    % \vspace{6pt}
    \centering
    % \vspace{-0.2cm}
    \setlength{\tabcolsep}{2pt} % Adjust the space between columns
    \setlength{\aboverulesep}{0.3pt}
    \setlength{\belowrulesep}{0.3pt}
    % \small
    \fontsize{8pt}{12pt}\selectfont % Set font size and line spacing
    \resizebox{0.49\textwidth}{!}{\begin{tabular}{c | c c |  c c c c c  }
    \toprule
    & \multicolumn{2}{c|}{Design Choices} & \multicolumn{5}{c}{Closed-loop metrics ($\uparrow$)}  \\
    \midrule
    Model Type & \makecell{Random\\Sample}   & \makecell{Guide\\Reward}   &  CLS-NR &  S-CR  & S-Area  & S-PR & S-Comfort   \\
    \midrule
    \multirow{2}{*}{\makecell{Vanilla}} 
        & {\cmark}   & Progress       &  67.56 $\pm$ 0.38 & 90.97 $\pm$ 0.78 & 94.64 $\pm$ 1.72 & 72.17 $\pm$ 0.21 & 64.21 $\pm$ 1.29    \\
        & {\cmark}& DE            &  86.89 $\pm$ 0.28 & \underline{97.34 $\pm$ 0.37} & 96.36 $\pm$ 0.18 & 89.90 $\pm$ 0.11 & \textbf{94.03 $\pm$ 0.65}    \\
        \midrule
    \multirow{2}{*}{\makecell{Consistent}}
        & {\xmark}    & {FE}       &  \underline{88.14} & 96.86  & \underline{98.43}  & \underline{91.39}  & 73.73    \\
        & {\xmark}    & {DE}       &  \textbf{94.07} & \textbf{99.22}  & \textbf{99.22}  & \textbf{95.06}  & 91.09    \\
    \bottomrule
    \end{tabular}}
    % \vspace{-0.3cm}
    \caption{Comparison of vanilla and consistent auto-regressive frameworks with different guide reward design. Experimental results are based on the Test14-Random non-reactive benchmark.}
    \label{table:supple-abla-rl-reward}
    % \vspace{-0.3cm}
\end{table}

\begin{table}[t]
    % \vspace{6pt}
    \centering
    % % \vspace{-0.2cm}
    \setlength{\tabcolsep}{2pt} % Adjust the space between columns
    \setlength{\aboverulesep}{0.02pt}
    \setlength{\belowrulesep}{0.02pt}
    % % \small
    \fontsize{5pt}{6pt}\selectfont % Set font size and line spacing
    \resizebox{0.48\textwidth}{!}{\begin{tabular}{ c c |  c | c c }
    \toprule
    % \multicolumn{2}{c|}{Design Choices} & \multicolumn{1}{c|}{Closed-loop metrics ($\uparrow$)}  & \multicolumn{2}{c}{Consistent Ratio ($\uparrow$)}  \\
    % \midrule
    \makecell{Model} & \makecell{Loss}    &  CLS-NR ($\uparrow$)  & \makecell{Consistent Ratio\\Lat ($\uparrow$)}   & \makecell{Consistent Ratio\\Lon ($\uparrow$)}    \\
    \midrule
        Vanilla & RL          &  86.89 ± 0.28 & 20.00 ± 0.10 & 8.33 ± 0.00 \\
        PLUTO~\cite{cheng2024pluto}  & IL           &  91.92  &   62.45 &  41.80  \\
        Consistent & IL       &  \underline{93.41} & \underline{68.26} & \underline{43.01} \\
        Consistent & RL       &  \textbf{94.07} & \textbf{79.58} & \textbf{43.03} \\
        \bottomrule
    \end{tabular}}
    % \vspace{-0.35cm}
    \caption{Comparison for consistency. Experimental results are based on the Test14-Random non-reactive benchmark.}
    \label{table:consistency}
    % \vspace{-0.8cm}
\end{table}

\begin{table}[t]
    % \vspace{6pt}
    \centering
    % \vspace{-0.2cm}
    \setlength{\tabcolsep}{3pt} % Adjust the space between columns
    \setlength{\aboverulesep}{0.3pt}
    \setlength{\belowrulesep}{0.3pt}
    % \small
    \fontsize{5pt}{6pt}\selectfont % Set font size and line spacing
    \resizebox{0.48\textwidth}{!}{\begin{tabular}{c |  c c c c c }
    \toprule
    & \multicolumn{5}{c}{Closed-loop metrics ($\uparrow$)}   \\
    \midrule
    \makecell{Transition Model}   &  CLS-NR &  S-CR  & S-Area  & S-PR & S-Comfort   \\
    \midrule
        Reactive &  91.03 & 96.92  & \textbf{99.23}  & 91.28  & 90.00  \\
        Non-reactive  &  \textbf{94.07} & \textbf{99.22}  & 99.22  & \textbf{95.06}  & \textbf{91.09}  \\
    \bottomrule
    \end{tabular}}
    % \vspace{-0.3cm}
    \caption{Comparison of the usage of reactive and non-reactive transition models. Experimental results are based on the Test14-Random non-reactive benchmark.}
    \label{table:supple-abla-rl-tm}
    % \vspace{-0.3cm}
\end{table}

%% file: blocks/figure_time_horizon.tex
\begin{figure}[t]
    \begin{center}
    \includegraphics[width=0.46\textwidth]{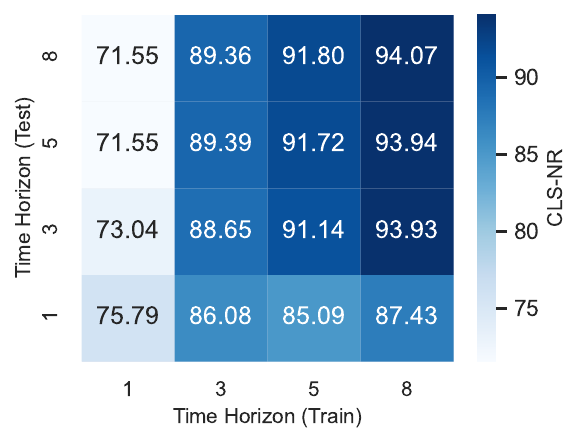}
    %%% trim={<left> <lower> <right> <upper>}
    \end{center}
    \vspace{-0.5cm}
    \caption{Performance of different training time horizons under different testing time horizons. The value in each cell is the CLS-NR metric on the Test14-Random non-reactive benchmark.
    }
    % \vspace{-0.6cm}
    \label{figure:supple-time-horizon}
\end{figure}

%% file: blocks/figure_vis2.tex
\begin{figure*}[ht]
    \begin{center}
    \includegraphics[width=0.98\textwidth]{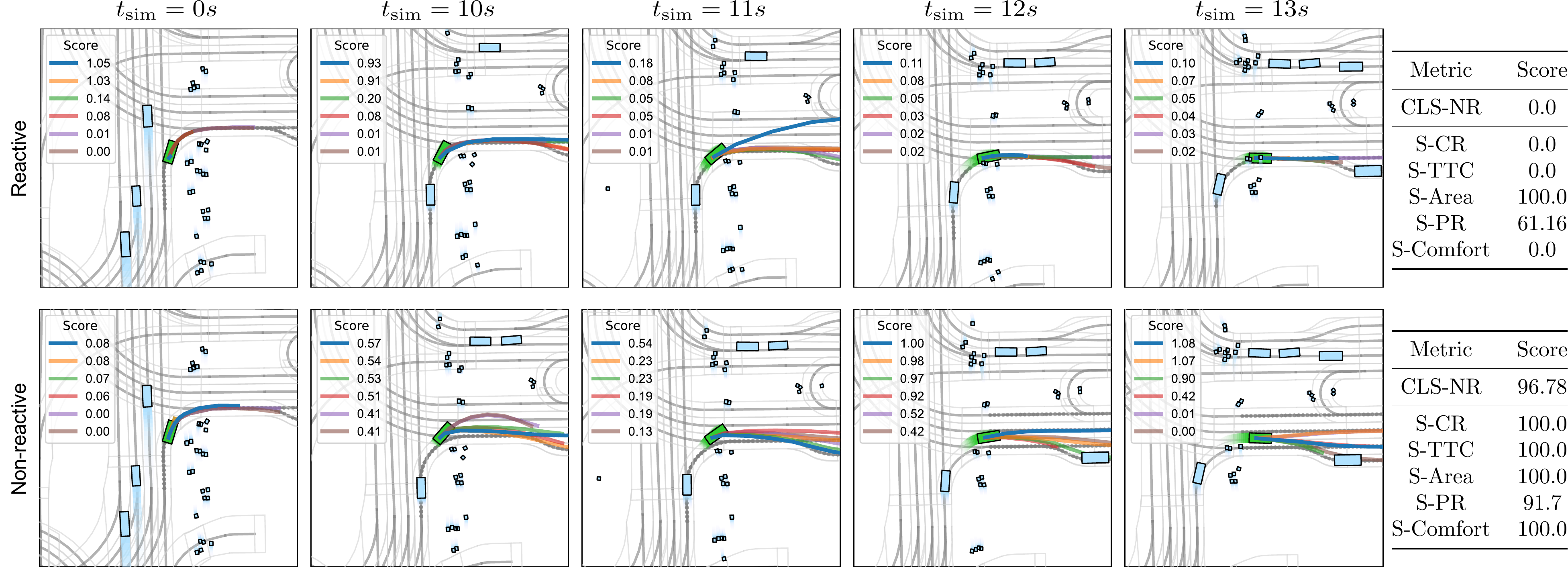}
    %%% trim={<left> <lower> <right> <upper>}
    \end{center}
    \vspace{-0.7cm}
    \caption{Qualitative comparison of using reactive and non-reactive transition model in non-reactive environments. The scenario is annotated as \texttt{waiting\_for\_pedestrian\_to\_cross}. In each frame shot, ego vehicle is marked as {\color{LimeGreen} green}. Traffic agents are marked as {\color{DeepSkyBlue} sky blue}. Lineplot with {\color{Blue} blue} is the ego planned trajectory.
    }
    % \vspace{-0.5cm}
    \label{figure:vis2}
\end{figure*}

%% file: blocks/table_supple_diff_loss.tex
\begin{table*}[t]
    % \vspace{6pt}
    \centering
    % \vspace{-0.2cm}
    \setlength{\tabcolsep}{3pt} % Adjust the space between columns
    \setlength{\aboverulesep}{0.3pt}
    \setlength{\belowrulesep}{0.3pt}
    % \small
    \fontsize{5pt}{6pt}\selectfont % Set font size and line spacing
    \resizebox{0.98\textwidth}{!}{\begin{tabular}{c | c c c |  c c c c c | c c }
    \toprule
    & \multicolumn{3}{c|}{Supervision Signals} & \multicolumn{5}{c|}{Closed-loop metrics ($\uparrow$)}  &  \multicolumn{2}{c}{Open-loop metrics ($\downarrow$)}  \\
    \midrule
    \makecell{Loss\\Type} & DE  & \makecell{Col}   & \makecell{Area}   &  CLS-NR &  S-CR  & S-Area  & S-PR & S-Comfort   & \makecell{Col\\Mean [Min, Max]} & \makecell{Area\\Mean [Min, Max]}  \\
    \midrule
    \multirow{4}{*}{\makecell{IL}} 
        & {\cmark} & {\xmark}   & {\xmark}      &  93.41 & 98.85  & \underline{98.85}  & 93.87  & \textbf{96.15}   &  0.17 [\textbf{0.00}, 0.47]  &  0.09 [\textbf{0.00}, 0.40]  \\
        & {\cmark} & {\cmark}   & {\xmark}      &  \underline{93.67} & \textbf{99.23}  & \underline{98.85}  & \underline{94.63}  & 94.23    &  0.16 [\textbf{0.00}, \underline{0.43}]  &  \underline{0.07} [\textbf{0.00}, \underline{0.27}]  \\ 
        & {\cmark} & {\xmark}   & {\cmark}      &  93.12 & 98.46  & 98.84  & 92.88  & 94.21    &  \underline{0.15} [\textbf{0.00}, 0.44]  &  0.08  [\textbf{0.00}, 0.30]  \\ 
        & {\cmark} & {\cmark}   & {\cmark}      &  93.32 & 98.46  & 98.46  & 94.05  & \underline{95.77}   &  \underline{0.15} [\textbf{0.00}, \underline{0.43}]  &  0.09 [\textbf{0.00}, 0.39]  \\ 
    \midrule
    \multirow{2}{*}{\makecell{RL}} 
        & {\cmark} & {\xmark}    & {\xmark}     &  90.44 & 97.49  & 96.91  & 93.33  & 90.73   &  0.17  [\textbf{0.00}, 0.49]  &  0.14 [\textbf{0.00}, 0.51]  \\
        & {\cmark} & {\cmark}    & {\cmark}     &  \textbf{94.07} & \underline{99.22}  & \textbf{99.22}  & \textbf{95.06}  & 91.09   &  \textbf{0.12}  [\textbf{0.00}, \textbf{0.39}]  &  \textbf{0.05} [\textbf{0.00}, \textbf{0.22}]  \\
        \bottomrule
    \end{tabular}}
    % \vspace{-0.3cm}
    \caption{Comparison with different loss types and supervision signals. Closed-loop results are based on the Test14-Random non-reactive benchmark. Open-loop results are on validation set.}
    \label{table:supple-abla-diff-loss}
    % \vspace{-0.3cm}
\end{table*}

\begin{table*}[t]
    % \vspace{6pt}
    \centering
    % \vspace{-0.2cm}
    \setlength{\tabcolsep}{3pt} % Adjust the space between columns
    \setlength{\aboverulesep}{0.3pt}
    \setlength{\belowrulesep}{0.3pt}
    % \small
    \fontsize{5pt}{6pt}\selectfont % Set font size and line spacing
    \resizebox{0.98\textwidth}{!}{\begin{tabular}{c | c c c c c c |  c c c c c}
    \toprule
     & \multicolumn{6}{c|}{Design Choices} & \multicolumn{5}{c}{Closed-loop metrics ($\uparrow$)}   \\
    \midrule
    \makecell{Loss\\Type} & \makecell{Model\\Type} & \makecell{Mode\\Type} & \makecell{Mode\\Dropout}   &  \makecell{Scorer\\Side Task}  &  \makecell{Ego-history\\Dropout} & \makecell{Backbone\\Sharing}   &  CLS-NR &  S-CR  & S-Area  & S-PR & S-Comfort \\
    \midrule
    \multirow{3}{*}{\makecell{IL}}
        & Vanilla & -  & {-}    & {-}   & {\cmark}   & {-}  &  86.48 & 97.09  & 97.29  & 88.05  & 94.19   \\
        & Consistent & Lon  & {\xmark}    & {\cmark}   & {\cmark}   & {\cmark}   &  88.79 & 96.67  & 96.08  & 89.63  & \underline{94.90} \\
        & Consistent & Lon-Lat & {\cmark}    & {\cmark}  & {\cmark}   & {\cmark}     &  \underline{93.41}             & \underline{98.85}  & \underline{98.85}                & \underline{93.87}                & \textbf{96.15}         \\ 
    \midrule
    \multirow{3}{*}{\makecell{RL}}
        & Vanilla & -  & {-}    & {-}   & {\xmark}   & {-}  &  85.56 & 97.27  & 95.70  & 89.17  & 93.36 \\
        & Consistent & Lon  & {\xmark}    & {\cmark}   & {\xmark}   & {\xmark}  &  90.57 & 97.30  & 97.68  & 92.20  & 94.59 \\
        & Consistent & Lon-Lat & {\cmark}    & {\cmark}    & {\xmark}   & {\xmark}   &  \textbf{94.07}    & \textbf{99.22}     & \textbf{99.22}    & \textbf{95.06}    & 91.09        \\ 
        \bottomrule
    \end{tabular}}
    % \vspace{-0.3cm}
    \caption{Effect of different mode representations. Experimental results are based on the Test14-Random non-reactive benchmark.}
    \label{table:supple-abla-mode}
    % \vspace{-0.3cm}
\end{table*}

%% file: blocks/figure_diff_loss.tex
\begin{figure}[t]
    \begin{center}
    \includegraphics[width=0.4\textwidth]{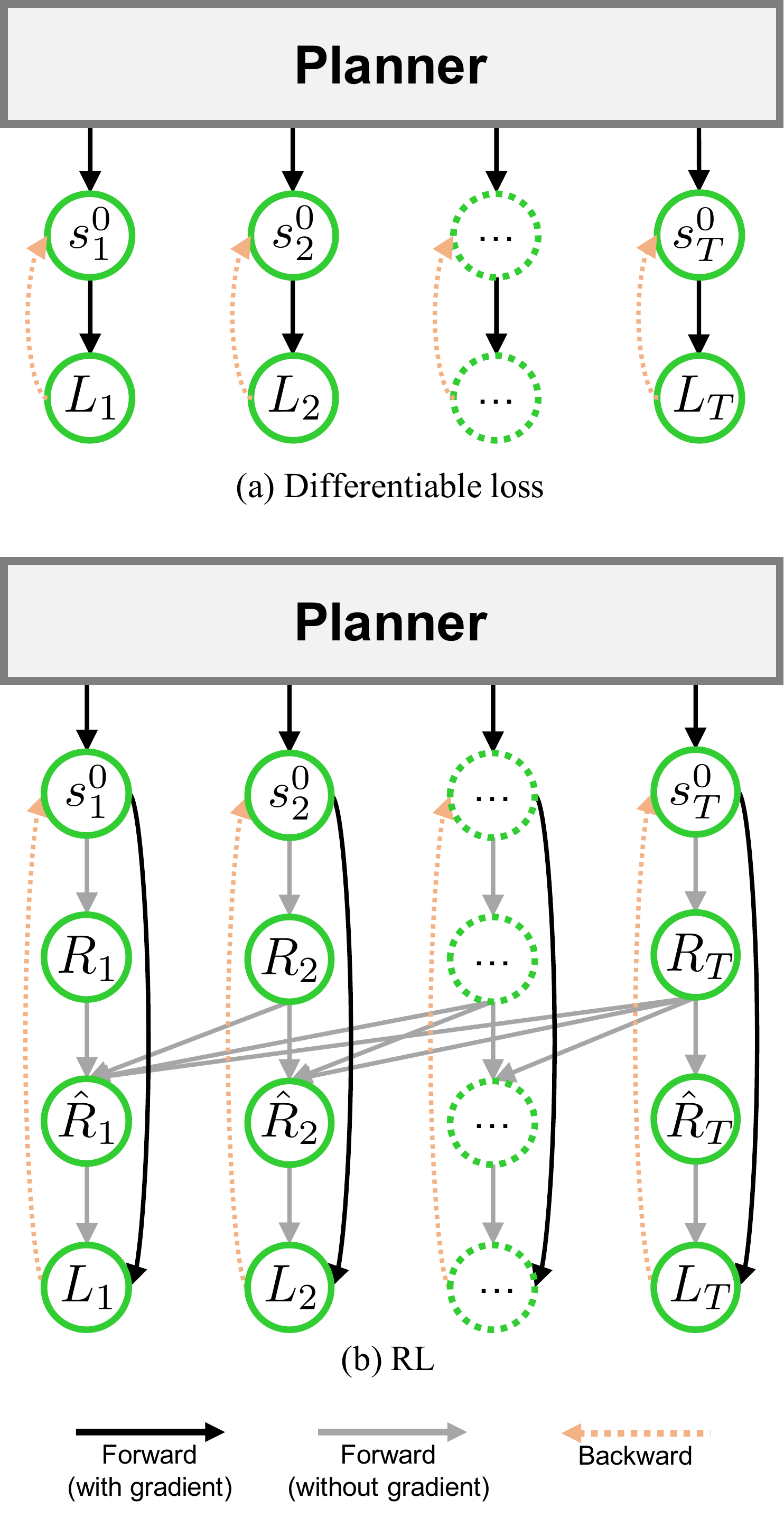}
    %%% trim={<left> <lower> <right> <upper>}
    \end{center}
    \vspace{-0.3cm}
    \caption{The computational graph of differentiable loss (a) and RL (b) framework for optimizing same metrics such as displacement errors, collision avoidance, and adherence to drivable area.}
    % \vspace{-0.5cm}
    \label{figure:diff-loss}
\end{figure}